\documentclass[]{llncs}

\usepackage{amssymb}
\usepackage{latexsym}

\usepackage{url}
\usepackage{xcolor}
\definecolor{newcolor}{rgb}{.8,.349,.1}

\usepackage{graphicx}

\usepackage{algorithm}
\usepackage{algorithmicx}
\usepackage{algpseudocode}
\usepackage{CJKutf8}
\usepackage{amsmath}
\usepackage{amssymb}

\begin{document}
\title{The Stroke Correspondence Problem, Revisited}


\author{Dominik Klein} 
\institute{}
\maketitle

\begin{abstract}
We revisit the stroke correspondence problem~\cite{WSNMO95,WSNMO96}. We optimize 
this algorithm by 1) evaluating suitable
preprocessing (normalization) methods 
2) extending the algorithm with an additional
distance measure to handle Hiragana,
Katakana and Kanji characters with a low number
of strokes and c) simplify the stroke linking
algorithms. Our contributions are implemented
in the free, open-source library \textsf{ctegaki}
and in the demo-tool \textsf{jtegaki}.\end{abstract}






\section{Introduction}

Revisiting the \emph{stroke correspondence problem} is motivated by developing a robust, freely available, open-source
Japanese on-line handwriting recognition engine, especially considering the current advance of touch-driven mobile devices. For handwriting recognition, 
learning based methods have largely replaced approaches based on template matching. Popular methods are directional feature extraction~\cite{KYHHMTM92} and classification by MQDF~\cite{KTTM87}, or approaches based on neural networks~\cite{LYWW11a}. In order to a achieve satisfactory performance, learning based methods however require lots of training samples for each character~\cite{RGB02}. Several large databases with training samples exist, cf.~\cite{SYY85,NM04,LYWW11b}, or the JEITA-HP database. To the author's best knowledge, none of them is available under a free license that would permit use in free/libre open-source software. On the other hand, generating such a database from scratch is no easy task; for example~\cite{NM04} write -- despite being full-time academic researchers -- that they  \textit{spent 4 years compiling [...] databases}, and such a monumental task is difficult to achieve within an open-source project. This is probably one main reason why none of the above mentioned high-performing recognition approaches have been implemented and released as open source.

It is very important for a recognition engine to be able to cope with stroke-order and stroke-number variations.  
While in theory the number and the order of strokes is 
uniquely defined for each kanji character, in 
practice they vary widely 
w.r.t.\ individual writing styles. Native speakers for example
tend to combine several strokes into one in order to write 
faster (cursive style), and foreign leaners of the language
often do not know the correct stroke order. A recognition 
algorithm must thus be able to recognize kanji
independent on stroke-order and stroke-number.

The only two open-source Japanese online character recognition engines
the author is aware of - \textsc{zinnia}\footnote{\url{http://www.zinnia.sourceforge.net}} and \textsc{wagomu}\footnote{\url{http://tegaki.org}} use learning based approaches and employ a rather small training data set of one sample per character. Therefore they cannot cope with stroke-number or stroke-order variations very well.

In this paper, a simple approach of directly linking a written character to a reference template pattern~\cite{WSNMO96} is revisited. Recognition is then performed by
directly comparing the reference template with the
input pattern, and the ability to cope with stroke-order and stroke-number variations is inherent in the comparison.
Such an approach has the big advantage, that no training data is needed, and one reference template for each character suffices. \cite{S02} used a similar approach using directional features for reference templates, resulting in a very high recognition rate. This comes at the cost of stroke-number free recognition, and a sophisticated but complex data-structure and search algorithm. 
In this paper the original approach~\cite{WSNMO96} is extended by
\begin{enumerate}
\item evaluating which image normalization techniques is  the overall performance.
\item introducing a distance measure using directional features akin to~\cite{S02}. This increases recognition performance of Hiragana, Katakana and Kanji with a low number of strokes.
\item Last, a simple-to-implement yet well-performing stroke linking algorithm is introduced, replacing the two complementing algorithms defined by~\cite{WSNMO96}.
\end{enumerate}
All techniques have been implemented in the open-source library \textsc{ctegaki}, which is available under a free BSD-style license at
\begin{center}
\url{https://github.com/asdfjkl/ctegaki-lib}
\end{center}
Moreover a JavaScript-based demo application is available at
\begin{center}
\url{https://asdfjkl.github.io/kanjicanvas}
\end{center}
This paper is structured as follows. In Section~\ref{sec:stroke_correspondence} the \emph{stroke correspondence problem} is reviewed and distance measures are introduced. Suitable normalization methods are discussed in Section~\ref{sec:normalization}. Two complementing algorithms were used by~\cite{WSNMO96} for stroke linking. A new algorithm is introduced in Section~\ref{sec:stroke_linking}. The overall recognition algorithm is presented in Section~\ref{sec:recognition}. The effect of various normalization algorithms, the newly introduced directional distance measure, and the new stroke linking algorithm, as well as overall performance compared to the open source recognition library \textsc{Zinnia}, is empirically evaluated in Section~\ref{sec:experiments}.
Last, the presentation is concluded in Section~\ref{sec5}. 

\section{The Stroke Correspondence Problem}\label{sec:stroke_correspondence}

A kanji character $k$ is defined as a set of strokes $k = {s_1,\ldots,s_n}$. This set is ordered w.r.t.\ the index of the strokes which represent the order in which the strokes were drawn. A strokes $s$ is a set of points $s=\{s_1,\ldots,s_k\}$, again ordered by their indices. A point $(x,y)$ is a 2D-coordinate.
Let $k_1$ be a kanji with $n$ strokes $s_1,\ldots,s_n$, and $k_2$ be a kanji with $m$ strokes $t_1,\ldots,t_m$. In the following we assume w.l.o.g. $n \geq m$.
Given two sets of strokes $S$ and $T$, a \emph{stroke distance function} is a function $d : S \times T \to \mathbb{R}$.

Given two kanji with the same number of strokes and a stroke distance function, the \emph{$n-n$ stroke correspondence problem} is the optimization problem to find a mapping between the strokes that minimizes the overall distance.
\begin{definition}[$n-n$ Stroke Correspondence Problem]
Let
$k_1 = \{s_1,\ldots,s_n\}$ and $k_2$ be two kanji both of length $n$, and $d$ a stroke distance function.
The \emph{$n-n$ stroke correspondence problem} is the problem to find a bijective function $f : k_1 \to k_2$, that minimizes $
\sum_{1 \leq i \leq n} d(s_i,f(s_i))$.
\end{definition}
The general idea for recognition is to define one template for each kanji. Given an input kanji that should be recognized, a minimizing function is computed for each pair of the input kanji and one template. Then the overall distance between the input kanji and the template is calculated w.r.t.\ the minimizing function, and the result of the recognition is the template with the smallest distance. Since the minimizing function does not incorporate any stroke order, this recognition approach is completely independent of the stroke order.

Two problems arise with this approach. First, given a kanji and a template with $n$ strokes, the cost of finding a minimizing function $f$ is in $\mathcal{O}(n!)$. Fast approximation algorithms for $f$ are subject of Section~\ref{sec:stroke_linking}.
Second, the definition can only cope with input kanji and templates that have the same number of strokes. To deal with kanji of arbitrary numbers of strokes, the definition is extended. Given two strokes $s_1$ and $s_2$, the \emph{concatenated stroke} of $s_1$ and $s_2$ is defined as the union $s_1 \cup s_2$, where the order of points is extended such that all points of $s_1$ are smaller than those of $s_2$, i.e.\ the points of $s_2$ are appended to $s_1$. Concatenation is extended in the obvious way to arbitrary numbers of strokes. Given $\{s_1, \ldots, s_k\}$, the concatenation of all strokes is denoted by $\mathsf{conc}(\{s_1, \ldots, s_k\})$. In the next definition, only surjectivity is required for $f$:
\begin{definition}[$n-m$ Stroke Correspondence Problem]
Suppose two kanji $k_1=\{s_1,\ldots,s_n\}$ and $k_2=\{t_1,\ldots,t_m\}$ are given, and let $d$ be a stroke distance function. The \emph{$n-m$ stroke correspondence problem} is the problem to find an surjective function $f : k_1 \to k_2$, that minimizes 
$
\sum_{1 \leq i \leq m} d(\mathsf{conc}(\{s \mid f(s) = t_i\}), t_i).
$
\end{definition}
This optimization problem depends on the definition of the stroke distance function. These are subject of the next section.
\subsection{Stroke Distance Functions}
Three distance functions were originally introduced by \cite{WSNMO96}.
\begin{definition}[Endpoint Distance]
Suppose two strokes $s_1=\{(x_1,y_1),\ldots,(x_n,y_n)\}$ and $s_2 = \{(u_1,v_1),\ldots,(u_m,v_m)\}$ are given. The \emph{endpoint distance} is defined as
\[
d_{\mathsf{ep}}(s_1,s_2) = |x_1 - u_1| + |y_1 - v_1| + |x_n - u_m| + |y_n - v_m|.
\]
\end{definition}
In the original definition, the calculation used division by some constant. This is omitted here. Since the computation of the endpoint distance is quite efficient, it can be used for a quick coarse classification to find suitable candidates. For fine classification, the next two distance measures were introduced. In the next definition, it is assumed that $m,n \in \mathbb{Q}$. Hence, division results in a real number.
\begin{definition}[Initial Stroke Distance]
Suppose two strokes $s_1=\{(x_1,y_1),\ldots,(x_n,y_n)\}$ and $s_2 = \{(u_1,v_1),\ldots,(u_m,v_m)\}$. Let $n \geq m$. The \emph{initial stroke distance} is defined as
\[
d_{\mathsf{in}}(s_1,s_2) = \frac{n}{m} \; \cdot \sum_{i=1}^{m} 
|x_i - u_i| + |y_i - v_i|
\]
\end{definition}
The next distance measure is the most precise one.
\begin{definition}[Whole-Whole Stroke Distance]
Suppose $s_1=\{(x_1,y_1),\ldots,(x_n,y_n)\}$ and $s_2 = \{(u_1,v_1),\ldots,(u_m,v_m)\}$. Let $n \geq m$. The \emph{whole-whole stroke distance} is defined as
\[
d_{\mathsf{ww}}(s_1,s_2) = 
\frac{1}{m} \cdot
\sum_{i=1}^{m} 
|x_i - u_{j(i)}| + |y_i - v_{j(i)}|
\]
Here $j(i) = \left(\frac{n-1}{m-1} \cdot (i-1)\right)+1$, integer division is used for the calculation of $i(j)$,
and the division $1/m$ for $d_{\mathsf{ww}}$ is in $\mathbb{Q}$.
\end{definition}
None of these stroke distances emphasize the actual stroke \emph{direction}. \cite{S02} used directional features  for stroke order independent (but not stroke number independent) recognition. Directional features are especially important when recognizing kanji with low number of strokes, or hiragana and katakana characters. Consider for example 
\begin{CJK}{UTF8}{min}し\end{CJK}
and 
\begin{CJK}{UTF8}{min}乙\end{CJK}.
If the stroke of these characters is represented by only a few points, all above mentioned stroke distances will result in a low overall distance, and hence distinguishing these characters is difficult. On the other hand, the two characters have very distinct directional features. This motivates the next definition, which is new compared to those introduced by \cite{WSNMO96}.
\begin{definition}[Directional Stroke Distance]
Let $s_1=\{(x_1,y_1),\ldots,(x_n,y_n)\}$ and $s_2 = \{(u_1,v_1),\ldots,(u_m,v_m)\}$, and let $n \geq m$. The \emph{directional stroke distance} is defined as
\begin{align*}
d_{\mathsf{dd}}(s_1,s_2) = \frac{1}{m} 
\cdot \sum_{i=2}^{m} 
&~\Big(|(x_i - x_{i-1}) - (u_{j(i)} - u_{j(i-1)})| \\
&+ |(y_i - y_{i-1}) - (v_{j(i)} - v_{j(i-1)})|\Big)
\end{align*}
where $j(i) = \left(\frac{n-1}{m-1} \cdot (i-1)\right)+1$ and $j(i)$ is computed with integer division, whereas the $1/m$ in $d_{\mathsf{dd}}$ is division in $\mathbb{Q}$. 
\end{definition}

\section{Normalization Methods}\label{sec:normalization}

Image Normalization is an important pre-processing step in character recognition. Given an image with height $h_1$ and width $w_1$ (i.e. a bounding box around the drawn character), image normalization projects this image to a an image with pre-defined height $h_2$ and width $w_2$. Throughout this section, we will consider a kanji as binary image (function) $f(x,y)$.

\begin{definition}[Linear Normalization]
Let $(x,y)$ be an input coordinate. Supposed $h_1$ and $w_1$ are the height and width of the input image, and $h_2$ and $w_2$ the height and width of the normalized image. The transformed coordinate $(x',y')$ is defined as:
$x' = x {w_2}/{w_1}$ and $y' = y{h_2}/{h_1}$.
\end{definition}
In the recognition approach of this paper, linear 
normalization is used only when the width and height of a character are very different. It then makes sense to keep the aspect ratio, and multiply only by the minimum of $w_2/w_1$
and $h_2/h_1$ (cf. Section~\ref{sec:recognition}).

\begin{figure}[t]
	\centering
  \includegraphics[width=0.8\textwidth]{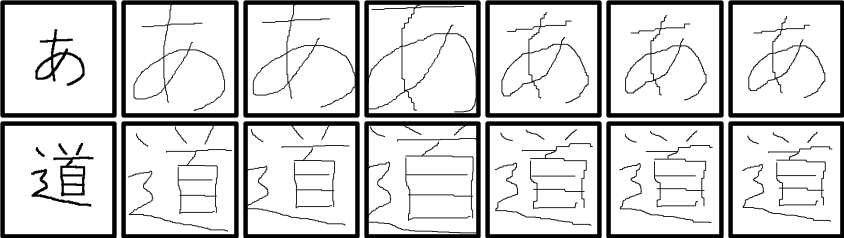}
	\caption{From left to right: Original Image, 
    Linear Normalization, Moment Normalization, Line Density Equalization, 
    Dot Density Equalization with $\alpha=2,3,4$.}
	\label{fig1}
\end{figure}

Moment normalization was originally formulated by \cite{C70}, but here the simplified variant introduced by \cite{LM05} is used.
Recall that given an image (function) $f(x,y)$, (raw) image moments are defined as $m_{ij} = \sum_{x} \sum_{y} x^i y^j f(x,y)$. Components of the centroid are $x'_c =
m_{10}/m_{00}$ and $y'_c = m_{01}/m_{00}$, and \emph{central moments} of the image are defined as 
$\mu_{pq} = \sum_{x}\sum_{y} (x - x'_c)^p \; (y - y'_c)^q \; f(x,y)$.
\begin{definition}[Moment Normalization]
Let $(x,y)$ be an input coordinate. Supposed $h_1$ and $w_1$ are the height and width of the input image, and $h_2$ and $w_2$ the heigth and width of the normalized image. Let $\delta_x = \sqrt{\mu_{20}}$ and $\delta_y = \sqrt{\mu_{02}}$. Distinct from other normalization methods, the bounding box of the input image is first reset to
$
\left[x_c - \frac{\delta_x}{2}, x_c + \frac{\delta_x}{2}\right],  
\left[y_c - \frac{\delta_y}{2}, y_c + \frac{\delta_y}{2}\right].
$
The transformed coordinate $(x',y')$ is defined as:
\[
x' = \frac{w_2}{\delta_x} (x - x_c) + x'_c, \quad\quad
y' = \frac{h_2}{\delta_y} (y - y_c) + y'_c
\]
\end{definition}
The effect on normalization techniques has been studied before especially in the context of directional feature extraction~\cite{LP94,LM05}. As can be seen in the latter, the introduction of advanced pseudo 2D normalization techniques improves recognition rates. However non-linear normalization~\cite{Y90,TT88} is a simple and fast technique that achieves a significant improvement over no or linear-normalization. The investigation is thus restricted here to the non-linear normalization techniques introduced by~\cite{Y90}. The next two methods rely on feature projection functions $H(x)$ and $V(y)$.
\begin{definition}[Dot Density Equalization]
Let $\alpha_H$ and $\alpha_V$ be given constants, and let $f(x,y)$ denote the input image, and $h_1,w_1,h_2,w_2$ the height and width of the input and normalized image. The projection functions are defined as:
$H(x) = \sum_{y=1}{h_1}f(x,y)+\alpha_H$ and
$V(y) = \sum_{x=1}{w_1}f(x,y)+\alpha_W$
The normalized image is defined as:
\begin{align}
x' &= \sum_{k=1}^{x}H(k) \cdot \frac{w_2}{\sum_{k=1}^{w_1}H(k)}, &
y' = \sum_{l=1}^{y}V(l) \cdot \frac{h_2}{\sum_{l=1}^{h_1}V(l)} 
\end{align}
\end{definition}
Line density equalization works similar as dot density equalization, but employs a more advanced density projection.
\begin{definition}[Line Density Equalization]
To define feature projections, first four types of edges $E$ are defined for a given point $(i,j)$. Below $\overline{f(i,j)}$ denotes the inverted pixel $f(i,j)$.
\begin{align*}
E_1 &= \mathsf{max} \{ i' \mid i' < i, f(i',j) \cdot \overline{f(i' + 1,j} = 1 \} \\
E_2 &= \mathsf{min} \{ i' \mid i' \geq i, f(i',j) \cdot \overline{f(i' + 1,j} = 1 \} \\
E_3 &= \mathsf{max} \{ i' \mid i' < i, \overline{f(i'-1,j)} \cdot f(i',j = 1 \} \\
E_4 &= \mathsf{min} \{ i' \mid i' \geq i, \overline{f(i'-1,j)} \cdot f(i',j = 1 \}
\end{align*}
Note that there is a case where $i'$ is not defined. Based on these four types of edges, horizontal and vertical \emph{line intervals} are defined. Let $w_1,h_1$ denote width and height of the input image.
$L_x$ is defined as 1) $4w_1$, if all $E_1,E_2,E_3,E_4$ are not defined, 2) $2w_1$, if only $E_1,E_3,$ are not defined, 3) $2w_1$, if only $E_2,E_4$ are not defined, 4)
$2w_1$, if only $E_1,E_4$ are not defined, 5) 
$E_4 - E_3$, if only $E_1$ is not defined, 6)
$E_2 - E_1$, if only $E_4$ is not defined, and finally $((E_2 - E_1) + (E_4 - E_3))/2$, otherwise.

The line interval $L_y$ for the vertical direction is defined analogously, where $h_1$ is taken instead of $w_1$, and edge definitions are changed w.r.t.\ vertical directions. Line densities are defined as:
$\rho(i,j) = \mathsf{max} (w_1/L_x, h_1/L_y)$
if $L_x$ + $L_y < 6w_1$, and $\rho(i,j) = 0$, otherwise.
Projection function are defined by 
\begin{align}
H(x) = \sum_{y=1}\rho(i,j), & &
V(y) = \sum_{x=1}\rho(i,j)
\end{align}
The actual mapping coordinates $(x',y')$ are then obtained by inserting
the projection functions of $(2)$ in equation $(1)$ of Definition 9.
\end{definition}

\section{Stroke Linking Algorithms}\label{sec:stroke_linking}

\begin{figure}[th]
	\centering
  \includegraphics[width=0.8\textwidth]{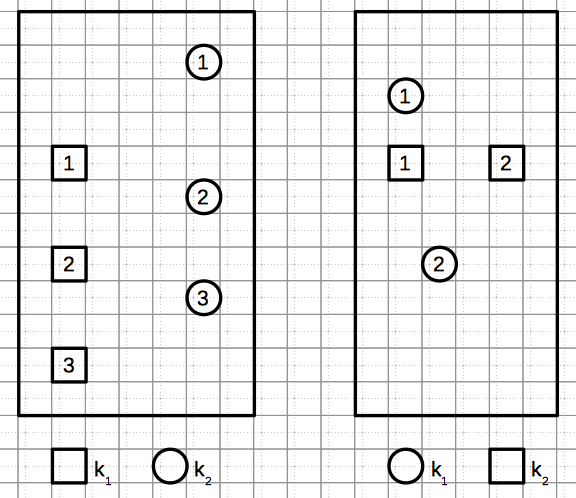}
	\caption{Examples of 2D-Point linking for
    the presented linking algorithms. DMD solves the left, but not 
    the right example, EMD
    solves the right, but not the left one, and iterative improvement solves both.
    }
    \label{fig:stroke_linking}
\end{figure}

The goal of a stroke linking algorithm is to find a (surjective) mapping function $f$ that meets the condition of Definition 2. In the original paper by \cite{WSNMO95}, the following approach was proposed: First, find a mapping from $n$ out $m$ strokes of kanji $k_1$ to the $n$ strokes of kanji $k_2$. Second, if $n \not = m$, the remaining strokes of $k_1$ are then concatenated with their preceding or following strokes. For the first step they propose two complementing algorithms, \emph{excessive mapping dissolution (EMD)} and \emph{deficient mapping dissolution (DMD)}. The motivation to introduce two algorithms is shown in Figure~\ref{fig:stroke_linking}. Considering a distance measure that simply computes the manhattan distance of two points, two $n/n$ mapping cases are depicted. For one case, EMD succeeds to find the optimal mapping whereas DMD fails, and in the other case, the reverse is true. 

Both from the perspective of run-time and implementation complexity, it is desirable to have only one single algorithm. A simple approach is just to use a greedy strategy, where, iterating over all $n$ strokes from $k_2$, given stroke $i$ from $k_1$, an unassigned stroke $j$ out of the $m$ stroke of $k_1$ is chosen for which the distance $d(i,j)$ is minimal. Manual inspection of such computed $n/n$ stroke maps revealed that in practice usually only very few excessive local minima cause a mapping that does not yield a global minimum. The typical cause of such local minima are strokes of very short length, which yield a very short local minimum with almost every nearby stroke, such as the first two strokes of
\begin{CJK}{UTF8}{min}字\end{CJK}.
This is the motivation to introduce the \emph{iterative improvement} algorithm. Technically this is one algorithm, but for the ease of presentation, the greedy initialization part as described above is shown separately as Algorithm 1, whereas the remaining part of the algorithm is shown as Algorithm 2.
Algorithm 2 loops a fixed set of $L$ times over the initialized stroke map, and in each loop switches the current assignment to another assigned stroke (line 9 to 17), or to another unassigned stroke (line 18 to 25), if the resulting distance can be improved by switching.

\begin{algorithm}[ht]
\caption{Greedy Initialization}\label{alg_init}
\begin{algorithmic}[1]
\Procedure{init}{kanjis $k_1,k_2$,distance function $d$}
\State $\text{map} \gets \text{new array($m$)}$
\For{$i \gets 1, n$}
\State $\text{map}[i] \gets -1$
\EndFor
\State $\text{free} \gets \{1,\ldots,n\}$
\For{$i \gets 1, m$}
\State $\text{min}_\text{dist} = \textsc{Maxvalue}$
\State $\text{min}_j = -1$
\For{$j \gets 1,n$}
\If{$j \in \text{free} \textbf{ and }d(k_1[i],k_2[j] < \text{min}_\text{dist}$}
\State $\text{min}_\text{dist} = d(k_1[j],k_2[i])$
\State $\text{min}_j = j$
\EndIf
\EndFor
\State $\text{free} \gets \text{free} \setminus \{\text{min}_j\}$
\State $\text{map[$\text{min}_j$]} = i$
\EndFor
\State \textbf{return} $map$
\EndProcedure
\end{algorithmic}
\end{algorithm}

\begin{algorithm}[ht]
\caption{$L$-Iterative Improvement}\label{alg_imp}
\begin{algorithmic}[1]
\Procedure{improvement}{kanji $k_1,k_2$,distance function $d$}
\State $\text{map} \gets \textsc{init}(k_1,k_2,d)$
\For{$1,\ldots,L$}
\For{$i \gets 1,\text{map.length}$}
\If{$\text{map}[i] \not = -1$}
\State $d_\text{ii} \gets d(k_1[i],k_2[\text{map[i]}])$
\For{$j \gets 1,\text{map.length}$}
\If{$\text{map}[i] \not = -1$}
\If{$\text{map}[j] \not = -1$}
\State $d_\text{jj} \gets d(k_1[j],k_2[\text{map[j]}])$
\State $d_\text{ij} \gets d(k_1[j],k_2[\text{map[i]}])$
\State $d_\text{ji} \gets d(k_1[i],k_2[\text{map[j]}])$
\If{$d_{\text{ji}} + d_{\text{ij}} < d_{\text{ii}} + d_{\text{jj}}$}
\State $\text{mapj} = \text{map}[j]$
\State $\text{map}[j] = \text{map}[i]$
\State $\text{map}[i] = \text{mapj}$
\State $d_\text{ii} = d_\text{ij}$
\EndIf
\Else
\State $d_\text{ij} \gets d(k_1[j],k_2[\text{map[i]}])$
\If{$d_{\text{ij}} < d_\text{ii}$}
\State $\text{map}[j] = \text{map}[i]$
\State $\text{map}[i] = -1$
\State $d_\text{ii} = d_\text{ij}$
\EndIf
\EndIf
\EndIf
\EndFor
\EndIf
\EndFor
\EndFor
\State \textbf{return} $map$
\EndProcedure
\end{algorithmic}
\end{algorithm}

Two examples are given in the style of \cite{WSNMO95}. Originally they illustrated the complementary nature of EMD and DMD, and here it is shown how iterative improvement succeeds in both cases.
For the ease of illustration, the examples consider strokes that contain only one point. Also for simplicity, the distance between two strokes is defined as the manhattan distance between the corresponding two points, and for that the example is slightly modified compared to the original example.

\begin{example}
Consider the left example of Figure~\ref{fig:stroke_linking}. Greedy initialization results in the following stroke map: 
$
m[1] = 2, \; m[2] = 3, \; m[3] = 1.
$
Here
the index of the array denotes a stroke of $k_1$, and the content is a stroke of $k_2$:
Next, Algorithm 2 iterates over strokes of $k_1$, and tests whether switching strokes decreases overall cost. First:
\begin{align*} 
d(1,m[1]) + d(2,m[2])  
\overset{?}{<}
d(1,m[2]) + d(2,m[1])
\end{align*}
Since $4 +4 < 7+5$, nothing is changed. The next comparison is:
\begin{align*}
d(1,m[1]) + d(3,m[3])
\overset{?}{<}
d(1,m[3]) + d(3,m[1]) 
\end{align*}
Since $4+12=16 \not < 6+8 = 14$, the assignments are switched and result in 
$
m[1] = 1, \; m[2] = 3, \; m[3] = 2.
$
It is not difficult to see that one additional iteration
will result in the optimal assignment
$m[1] = 1, \; m[2] = 2, \; m[3] = 3$.
\end{example}

\begin{example}
Consider the right example of Figure~\ref{fig:stroke_linking}.
Greedy initialization results in:
$
m[1] = 1, \; m[2]=2.
$
The first step of iterative improvement compares
\begin{align*}
d(1,m[1]) + d(2,m[2])
\overset{?}{<}
d(1,m[2]) + d(2,m[1])
\end{align*}
Since $3+4=7 \not < 4+1 = 5$, the optimal map
$m[1] = 2, \; m[2]=1$ is immediately found.
\end{example}

\section{Recognition Algorithm}\label{sec:recognition}

Putting all previous sections together and taking into account the original approach by \cite{WSNMO95}, the algorithm presented below is used for recognition. The input is an unknown kanji $k$. It is assumed that templates $t_1,t_2,\ldots$ exist, where one
templates corresponds to one unique kanji character.

\subsection{Recognition Algorithm}

\textbf{Interpolation:} The input kanji's resolution might not be dense enough. For example for a straight line, only the input and end coordinate might be specified. To get a higher resolution, intermediate points between each given point and its successor are computed using Bresenham's line algorithm.

\textbf{Normalization:} The input kanji is normalized to an area of 256x256
pixels. To avoid excessive distortions for kanji characters that have a large difference w.r.t.\ length
and heigth (such as e.g.
\begin{CJK}{UTF8}{min}一\end{CJK}), linear normalization
is used whenever 3*heigth $>$ width or vice versa.
Otherwise one of the following is used:
linear normalization, moment normalization,
normalization by line density equalization, or normalization by dot density equalization with $\alpha = 2,3$ or $4$.

\textbf{Interpolation:}
Normalization can decrease the resolution, and thus again intermediate points are interpolated using Bresenham's line algorithm.

\textbf{Feature Point Extraction:} For each stroke, feature points are extracted. The initial and end point of a stroke are always extracted. Intermediate points are extracted in a way such that the euclidian distance between two consecutive points is close to a fixed global value. In the implementation, a fixed value of 20.0 is used. The result of this step is a kanji $k'$ with the same number of strokes as the input kanji, but each stroke has a much lower number of (extracted feature) points.

\textbf{Coarse Classification:} 
It is assumed that each template has been preprocessed in the same way as described above. Next coarse classification finds a predefined set (in our implementation: 100) of candidates, in the following way:

For each template $t$, a stroke map of $k'$ and $t$ is computed. First, the stroke map is initialized by Algorithm 1, and improved in $L$ steps by Algorithm 2. In both cases endpoint distance is used. If both $k'$ and $t$ have the same number of strokes, the overall endpoint distance between $k'$ and $t$ is calculated using the computed stroke map. Otherwise, the stroke map $m$ is completed in the following way: 
\begin{enumerate}
\item Let $(i,m[i])$ be the smallest $i$ where $m[i] \not = -1$. Then all strokes $j < i$ are mapped to $m[i]$.
\item Let $(i,m[i])$ be the largest $i$ such that $m[i] \not = -1$ and all $j > i$ have $m[j] = -1$. Then all strokes $j>i$ are mapped to $m[i]$.
\item Suppose that there are $i,j$ with $j>i$, $m[i] \not= -1$, $m[j] \not= -1$, and for all $k$ with $i<k<j$ there is $m[k] = -1$. A split point $l$ can be chosen among all $k$ in the following way: $m[k]$ is set to $m[i]$ for all $k \leq l$, and $m[k]$ is set to $m[j]$ for all $k>l$. The split point $l$ is chosen in a way such that the overall computed distance is minimal. 
\end{enumerate}
The result of the coarse classification are those 100 templates, that yield the smallest overall $n-m$ stroke correspondence between $k'$ and the template with the above computed stroke map using the endpoint distance measure. 

\textbf{Fine Classification:}
During fine classification, ten templates are chosen among those 100 that are selected by coarse classification. First a stroke map between $k'$ and each template is computed using Algorithm 1 and 2 with the initial distance measure. If template and $k'$ do not possess the same number of strokes, the stroke map is completed as in coarse classification. However instead of using the endpoint distance measure, the following approach is used: 
\begin{enumerate}
\item
If the input $k'$ has less than a predefined number $S$ of strokes, then the stroke map is completed using the directional distance measure.
\item
Otherwise, the whole-whole distance measure is used. The predefined number is chosen empirically, cf. Table~\ref{tbl:iterative}.
\end{enumerate}
Finally a weight for each template is computed.
Let $\gamma = \gamma_1, \ldots, \gamma_m$.
Below, $W_{k',t,\gamma,d}$ is written for 
$
\sum_{1 \leq i \leq m} \gamma_i \cdot d(\mathsf{conc}(\{s \mid f(s) = t_i\}, t_i).
$
Using the above computed stroke map, the following weight $W$ is computed:
\[
W = 
\begin{cases}
\frac{1}{\text{min}(n,m)} W_{k',t,\gamma,d_{\mathsf{ww}}}
&\text{if $k'$ has less than $S$ strokes} \\[1em]
\frac{1}{\text{min}(n,m)} W_{k',t,\gamma,d_{\mathsf{dd}}}
&\text{if $k'$ has less than $S$ strokes}
\end{cases}
\]
Here, for two strokes $s_i, t_i$ containing $l$ and $o$ points,  $\gamma_i = \text{max}(l,o) / \text{min}(l,o)$, if $s_i$ or $t_i$ resulted from concatenation, and $\gamma = 1.0$, otherwise. The result of the fine classification are those ten templates, that yield the lowest weights $W$.

\begin{table}[t]
\centering
\caption{Comparison of different normalization methods.}
\resizebox{\columnwidth}{!}{%
\begin{tabular}{l*{6}{c}r}
\hline\\ [-1.5ex]
& Linear~  & Moment & DotDensity& DotDensity& DotDensity & LineDensity  \\
     &      &    & $\alpha=2$ & $\alpha=3$ &$\alpha=4~$\\
\hline \\ [-1.5ex]
Kanji (79) &  &  &  &  &  &    \\
\; Top 1   &63  &62  &61  &62  &60  &60    \\
\; Top 5   &72  &73  &70  &71  &72  &70   \\
\; Top 10  &74  &75  &73  &73  &74  &71   \\
\\
Hiragana (46) &  &  &  &  &  &    \\
\; Top 1   &30  &37  &35  &33  &29  &26    \\
\; Top 5   &43  &44  &45  &45  &45  &49   \\
\; Top 10  &45  &45  &45  &45  &45  &32   \\
\\
Katakana (46) &  &  &  &  &  &    \\
\; Top 1   &32  &41  &32  &33  &32  &35    \\
\; Top 5   &46  &46  &46  &46  &46  &46   \\
\; Top 10  &46  &46  &46  &46  &46  &46   \\[1ex]
\hline\\ 
\end{tabular}
}
\label{tbl:comp_normalization}
\end{table}

\begin{table}[th]
\centering
\caption{Comparison of DMD, EMD, and iterative improvement.}
{%
\begin{tabular}{lc@{\hskip 1em}c@{\hskip 1em}c@{\hskip 2em}c@{\hskip 1em}c@{\hskip 1em}c@{\hskip 1em}c}
\hline\\ [-1.5ex]
&               \multicolumn{6}{c}{Stroke Linking Algorithm} \\
& \multicolumn{3}{c}{Kanji (79)} &\multicolumn{3}{c}{Timing (ms)}\\
& Top1  & Top5 & Top10& Min. & Max. & Avg.\\
\hline \\ [-1.5ex]
Original~\cite{WSNMO95}   &  &  &  &  &  & \\
\;DMD       & 62 & 73 & 74 & 5 & 207 & 16\\
\;EMD       & 59 & 68 & 71 & 4 & 135 & 13\\
\;combined  & 62 & 73 & 75 & 7 & 294 & 24\\
\\
\multicolumn{2}{l}{$L$-Iterative Improvement}  &  &  &  &  &  \\
\; (1,1)   & 59 & 68 & 73 & 4 & 137 & 16 \\
\; (1,2)   & 62 & 73 & 75 & 4 & 136 & 15 \\
\; (1,3)   & 62 & 73 & 75 & 4 & 142 & 18 \\
\\
\; (2,1)   & 59 & 69 & 73 & 5 & 148 & 19 \\
\; (2,2)   & 62 & 73 & 75 & 5 & 171 & 21 \\
\; (2,3)   & 62 & 73 & 75 & 5 & 171 & 22 \\
\\
\; (3,1)   & 59 & 69 & 73 & 6 & 168 & 25 \\
\; (3,2)   & 62 & 73 & 75 & 6 & 168 & 26 \\
\; (3,3)   & 62 & 73 & 75 & 6 & 176 & 28 \\[1ex]
\hline\\ 
\end{tabular}
}
\label{tbl:iterative}
\end{table}

\begin{table}[t]
\centering
\caption{Overall performance, and directional stroke distance measure.}
{%
\begin{tabular}{lc@{\hskip 2em}c@{\hskip 1em}c@{\hskip 1em}c@{\hskip 1em}c@{\hskip 1em}c@{\hskip 1em}c@{\hskip 1em}c}
\hline\\ [-1.5ex]
&                  & \multicolumn{5}{c}{Directional vs. Whole-Whole} \\
& \textsc{zinnia}  &S=0 & S=1& S=2& S=3 & S=4  &S=5\\
\hline \\ [-1.5ex]
Kanji (79) &  &  &  &  &  &   & \\
\; Top 1   &36  &63  &63  &63  &62  &61 & 61  \\
\; Top 5   &46  &76  &76  &76  &76  &76 & 74 \\
\; Top 10  &47  &76  &76  &76  &76  &76 & 74 \\
\\
Hiragana (46) &  &  &  &  &  &  &  \\
\; Top 1   &27  &35  &35  &34  &33  &33 & 33  \\
\; Top 5   &35  &42  &42  &43  &43  &45 & 45 \\
\; Top 10  &37  &44  &44  &46  &46  &46 & 46 \\
\\
Katakana (46) &  &  &  &  &  &  &  \\
\; Top 1   &34  &37  &37  &39  &44  &46  & 46 \\
\; Top 5   &41  &45  &45  &45  &46  &46  & 46\\
\; Top 10  &42  &45  &45  &45  &46  &46  & 46\\[1ex]
\hline\\ 
\end{tabular}
}
\label{tbl:comp_directional}
\end{table}
\section{Experiments}
\label{sec:experiments}
To evaluate the implemented recognition algorithm, 2264 templates were created. This set contains one template each for all \emph{jouyou kanji} (kanji characters for every day use defined by the Japanese Ministry of Education), as well as one template each for all hiragana and katakana characters. To test the recognition performance, a test set of 79 characters was prepared. These include 49 characters from the distribution of \textsc{Tegaki-Lab}\footnote{\url{https://github.com/cburgmer/tegaki/tree/master/tegaki-lab}}, and the remaining characters include various characters that were deliberately written in a sloppy semi-cursive style with stroke concatenations on the one hand, and on the other hand with mistakes in stroke order and style that a typical foreign learner of the language would make. Normalization experiments and stroke linking experiments were conducted with a graphical prototype application written in Java. The recognition part was then ported to pure C, and the remaining experiments were done with that C-Library. All tests were run on an Intel Core i5 @ 2.5 Ghz with 8 GB RAM running Mac OS X 10.9.4.

\subsubsection*{Normalization Methods}

For normalization, the recognition method described in Section~\ref{sec:normalization} was used. For fine classification, only the initial stroke distance (as the input to Algorithm 1 and 2) as well as whole-whole stroke distance (for computation of the weights) was used. The directional stroke distance was not utilized. The results are depicted in Table~\ref{tbl:comp_normalization}. Whereas in statistical classification using directional features non-linear normalization is known to significantly improve recognition performance~\cite{KYHHMTM92}, 
contrary to what was conjectured by~\cite{WSNMO95}, the recognition performance is actually not increased by non-linear normalization methods. In fact for kanji characters, simple linear normalization outperforms all non-linear normalization methods. On the other hand, when it comes to hiragana and katakana characters, which are more similar in style to western handwriting, it can be seen that moment normalization significantly outperforms all other methods.
The performance of non-linear normalization methods can be explained by the fact that they put more emphasis on and unify directional features at the cost of introducing distortions. Statistical classification methods that depend on directional features are not affected by this, but the distortions cause misrecognition for template matching. Thus
in the implemented library and in further experiments, moment normalization was used due to its good overall performance.

\subsection{Stroke Linking Algorithms}

Next the performance of $L$-iterative improvement was tested. Fine classification was conducted with the same distance measures as in the previous section.
However computation of the stroke map was done once with the algorithms by~\cite{WSNMO95}, i.e. only DMD, only EMD, and combined, and once with iterative improvement.
The results are depicted in Table~\ref{tbl:iterative}. 
Note that each combination is run twice, once for the computation of a stroke map during coarse classification and fine classification each. The first (second) number in brackets for iterative improvement denote what the value $L$ was set to during coarse (fine) classification.

As can be seen, DMD itself performs quite well but fails in some examples. EMD itself performs not very well, but complements DMD and improves overall recognition performance. As for $L$-iterative improvement, a combination of $L=(1,2)$ suffices to recognize all those characters that can be recognized by DMD \& EMD. But the average recognition time compared to the combination of DMD \& EMD is improved by $(1,2)$-iterative improvement  by more than one third, and for particular time intensive characters with a large number of strokes, the maximal recognition time is decreased by more than 50 percent.

\subsection{Overall Performance and Directional Stroke Distance}

Last, the recognition part of the Java prototype was ported to C. Due to different rounding results, especially w.r.t.\ to feature point extraction, slightly different recognition results compared to the Java prototype were obtained. Coarse classification was run with endpoint distance and $3$ iterative improvement, and fine classification was run also with $3$ iterative improvement but initial distance to get an $n-n$ strokemap, and the the stroke map was completed either using directional distance if the input character has less than $S$ strokes, or by using using directional distance, otherwise. Also, the performance was compared to the open-source online Japanese handwriting recognition library~\textsc{Zinnia}, which uses a learning based approach with support vector machines. \textsc{Zinnia} was trained with exactly the same templates, and the same test data was supplied for recognition. The results are depicted in Table~\ref{tbl:comp_directional}. As can be seen, a larger factor of $S$ improves recognition for hiragana and katakana characters, since they can be better distinguished using directional features. On the other hand, recognition performance of kanji characters decreases slightly. Nevertheless, the presented template based matching approach significantly outperforms the learning based approach of \textsc{Zinnia}, which performs poorly due to having only one training sample per character.

\section{Conclusion and Future Work}\label{sec5}

Ideally, a large set of training samples for Japanese on-line character recognition would be available under an open-source (BSD-style) license, and a learning based approach could then be used for open-source Japanese on-line handwriting recognition. Since this is not the case, and such a database is unlikely to appear in near future due to the significant cost and logistics required, the focus here was on a template based matching approach. By empirically studying different normalization methods, extending the original approach by a new distance measure, and simplifying and unifying the underlying stroke linking algorithms, a fast and highly accurate implementation was achieved. Moreover it could be shown that in the present test set, the template based approach significantly outperforms competing learning-based implementations.

\bibliographystyle{splncs04}
\bibliography{main}

\begin{thebibliography}{10}
\providecommand{\url}[1]{\texttt{#1}}
\providecommand{\urlprefix}{URL }
\providecommand{\doi}[1]{https://doi.org/#1}

\bibitem{C70}
Casey, R.: Moment normalization of handprinted characters. IBM Journal of
  Research and Development  \textbf{14}(5),  548--557 (Sep 1970)

\bibitem{KYHHMTM92}
Kawamura, A., Yura, K., Hayama, T., Hidai, Y., Minamikawa, T., Tanaka, A.,
  Masuda, S.: Online recognition of freely handwritten {J}apanese characters
  using directional feature densities. In: Pattern Recognition, 1992. Vol.II.
  Conference B: Pattern Recognition Methodology and Systems, Proceedings., 11th
  IAPR International Conference on. pp. 183--186 (Aug 1992)

\bibitem{KTTM87}
Kimura, F., Takashina, K., Tsuruoka, S., Miyake, Y.: Modified quadratic
  discriminant functions and the application to {C}hinese character
  recognition. Pattern Analysis and Machine Intelligence, IEEE Transactions on
  \textbf{PAMI-9}(1),  149--153 (Jan 1987)

\bibitem{LP94}
Lee, S.W., Park, J.S.: Nonlinear shape normalization methods for the
  recognition of large-set handwritten characters. Pattern Recognition
  \textbf{27}(7),  895 -- 902 (1994)

\bibitem{LM05}
Liu, C.L., Marukawa, K.: Pseudo two-dimensional shape normalization methods for
  handwritten {C}hinese character recognition. Pattern Recognition
  \textbf{38}(12),  2242 -- 2255 (2005)

\bibitem{LYWW11b}
Liu, C.L., Yin, F., Wang, D.H., Wang, Q.F.: {CASIA} online and offline
  {C}hinese handwriting databases. In: Proceedings of the 2011 International
  Conference on Document Analysis and Recognition. pp. 37--41. ICDAR '11 (2011)

\bibitem{LYWW11a}
Liu, C.L., Yin, F., Wang, Q.F., Wang, D.H.: {ICDAR 2011} {C}hinese handwriting
  recognition competition. In: Document Analysis and Recognition (ICDAR), 2011
  International Conference on. pp. 1464--1469 (Sept 2011)

\bibitem{NM04}
Nakagawa, M., Matsumoto, K.: Collection of on-line handwritten {J}apanese
  character pattern databases and their analyses. Int. J. Doc. Anal. Recognit.
  \textbf{7}(1),  69--81 (2004)

\bibitem{RGB02}
Rowley, H., Goyal, M., Bennett, J.: The effect of large training set sizes on
  online {J}apanese {K}anji and {E}nglish cursive recognizers. In: Frontiers in
  Handwriting Recognition, 2002. Proceedings. Eighth International Workshop on.
  pp. 36--40 (2002)

\bibitem{SYY85}
Saito, T., Yamada, H., Yamomoto, K.: On the data base {ETL9} of handprinted
  characters in {JIS} chinese characters and its analysis (in japanese).
  Transactions IECE Japan  \textbf{J68}(D(4)),  757--764 (1985)

\bibitem{S02}
Shin, J.p.: Optimal stroke-correspondence search method for on-line character
  recognition. Pattern Recogn. Lett.  \textbf{23}(5),  601--608 (Mar 2002)

\bibitem{TT88}
Tsukumo, J., Tanaka, H.: Classification of handprinted {C}hinese characters
  using nonlinear normalization and correlation methods. In: Pattern
  Recognition, 1988., 9th International Conference on. pp. 168--171 vol.1 (Nov
  1988)

\bibitem{WSNMO95}
Wakahara, T., Suzuki, A., Nakajima, N., Miyahara, S., Odaka, K.: On-line
  cursive kanji character recognition as stroke correspondence problem. In:
  Document Analysis and Recognition, 1995., Proceedings of the Third
  International Conference on. vol.~2, pp. 1059--1064 (Aug 1995)

\bibitem{WSNMO96}
Wakahara, T., Suzuki, A., Nakajima, N., Miyahara, S., Odaka, K.: Stroke-number
  and stroke-order free on-line kanji character recognition as one-to-one
  stroke correspondence problem. IEICE Transactions on Information and Systems
  \textbf{E79-D}(5),  529--534 (1996)

\bibitem{Y90}
Yamada, H., Yamamoto, K., Saito, T.: A nonlinear normalization method for
  handprinted kanji character recognition—line density equalization. Pattern
  Recognition  \textbf{23}(9),  1023 -- 1029 (1990)

\end{thebibliography}

\end{document}